\documentclass[11pt]{article}

\usepackage[T1]{fontenc}
\usepackage[utf8]{inputenc}
\usepackage{lmodern}

\usepackage[a4paper]{geometry}

\usepackage{graphicx}
\usepackage{amsmath,amssymb}
\usepackage{booktabs}
\usepackage{multirow}
\usepackage{makecell}
\usepackage{tabularx}
\usepackage{subcaption}
\usepackage{enumitem}
\usepackage{setspace}
\usepackage{listings}
\usepackage{float}
\usepackage{natbib}
\setcitestyle{round}
\usepackage[dvipsnames]{xcolor}
\usepackage[most]{tcolorbox}

\definecolor{promptblue}{RGB}{59,89,182}
\definecolor{mDarkTeal}{HTML}{23373B}
\definecolor{mLightTeal}{HTML}{2CB1BC}
\definecolor{mBg}{HTML}{F7F7F7}

\usepackage[colorlinks=true,linkcolor=blue,citecolor=blue,urlcolor=blue]{hyperref}

\newtcolorbox{bluebox}[1][]{
  enhanced,
  colback=mBg,
  colframe=mDarkTeal,
  coltitle=white,
  colbacktitle=mDarkTeal,
  title=#1,
  fonttitle=\bfseries\small,
  boxrule=0.8pt,
  top=4pt, bottom=4pt, left=5pt, right=5pt
}

\title{\textbf{OptiAgent: End-to-End Optimization Modeling via Multi-Agent Iterative Refinement}}

\author{
Adriana Laurindo Monteiro\thanks{Equal contribution as first authors. } \and
Nayse Fagundes\footnotemark[1] \and
Gabriel Mattos Langeloh \and
Gustavo de Oliveira Kanno \and
Priscila Louise Aguirre \and
Thiago Costa Rizuti da Rocha\thanks{Corresponding author.} \and
Victor Leme Beltran
\\[6pt]
 Instituto de Ciência e Tecnologia do Itaú, Brasil \\
\small
\begin{tabular}{c}
\texttt{mlaurindodrica@gmail.com}, 
\texttt{naysesfagundes@gmail.com} \\
\texttt{gabriel.langeloh@itau-unibanco.com.br},
\texttt{gustavo.kanno@itau-unibanco.com.br} \\
\texttt{priscila.aguirre@itau-unibanco.com.br}, 
\texttt{thiago.rizuti-rocha@itau-unibanco.com.br} \\
\texttt{victor.beltran@itau-unibanco.com.br}
\end{tabular}
}

\date{}

\begin{document}

\maketitle

\begin{abstract}
We propose OptiAgent, a multi-agent framework that, given a natural language description of an Operations Research problem, is able to output a solver-ready mathematical formulation as well as executable code. Our architecture prioritizes the mathematical modeling step, where dedicated agents extract structures, such as decision variables and constraints, enabling iterative self-correction. We introduce a novel multi-loop validation architecture with four specialized feedback mechanisms, each targeting a distinct failure mode such as misinterpretation, structural defects, mathematical inconsistencies, validation failures, and code errors. Alongside accuracy, our modular design improves the process of solving optimization problems by improving transparency, as each agent exposes its reasoning and feedback, making the full modeling process auditable. Our framework achieves state-of-the-art performance on 3 out of 4 benchmarks across LP, MILP, and Nonlinear Programming tasks, while remaining highly competitive on the remaining dataset.


\end{abstract}
\section{Introduction}
The increasing performance of Large Language Models (LLMs), whether it is caused by a higher number of parameters, or larger context windows or better reasoning capabilities has been accompanied with harder and even more complex tasks to Artificial Intelligence (AI). One example in this context is the use of LLMs to model optimization problems described in Natural Language (NL).

The ultimate goal is to use the reasoning capabilities of LLMs to identify the mathematical structure behind a NL description of an optimization problem. The model has to extract abstract entities of the problem: the decision variables, the feasible set structured by constraint functions and the objective function. The process of formulating a precise mathematical model, that is, to translate the natural language description, is highly costly: one has to be an Operations Research (OR) expert. This poses a huge barrier to industry applications, since solvers require precisely formulated mathematical models.

In this context, the use of LLMs as solvers or modeling assistants is a valuable tool as it can reduce the time and cost of the optimization modeling process. The research field of OR is concerned with a huge variety of classical optimization problems, ranging from Linear Programming (LP), Mixed-Integer Linear Programming (MILP), nonlinear, stochastic, convex and nonconvex problems to combinatorial problems. Although previous research has established the use of LLMs for formulating and solving optimization problems described by NL, a few challenges remain unsolved.  
%\red{Acho esse final um pouco desconexo considerando o próximo parágrafo... fico com vontade de saber quais são esses desafios. Encaixar melhor o parágrafo seguinte e fazer isso tudo conectar melhor com as contribuições.}

In particular, moving from research demonstrations to practical tools exposes several methodological and usability challenges. As discussed in \citep{ijcai2025p1192}, we can cite the cost of crafting a highly specialized dataset and the need of better evaluation methods, usually measured in solution accuracy (or optimality gap, solving time etc). A common blind spot in many proposed systems is generating executable code that may output the correct optimal value coming from an incorrect mathematical formulation. Besides that, a crucial property for real world applications is the need of transparency in the modeling process.

%\red{JÁ RESPONDE A COISA DO USUARIO?}
Thinking about expert users, who can infer the mathematical model from a natural-language description, being able to inspect an LLM’s decisions in a modular, step-by-step way is essential for debugging and refining generated OR models. The OR community increasingly emphasizes \citep{DEBOCK2024249,ijcai2025p1192} transparent and user-friendly modeling frameworks as a foundation for trust, reliability, and effective human oversight. As new LLM-based frameworks for automatic mathematical formulation of optimization problems are still under active development, we believe auditable model generation benefits both more experienced users and future framework developers, who can then focus their efforts on improving the main sources of errors in current systems.

\subsection{Contributions}
Our primary contribution lies in enhancing the explainability of the modeling process. To address this challenge, we introduce a modular model-agnostic multi-agent framework. Instead of treating optimization modeling as an end-to-end black-box process, our framework allows users to actively debug, refine, and interpret the generated models with clarity and confidence. %\red{JÁ RESPONDE A COISA DO USUARIO?} 
By doing so, we present a trustworthy modeling framework specifically designed to be accessible to non-expert users and to provide full access to modeling details and reasoning to more experienced users. Additionally, our experiments provide evidence that the proposed system is model-agnostic, as the architecture generalizes effectively across different LLM families.

The key technical contribution is a new feedback-driven architecture that enables automated error detection and correction without requiring additional user intervention. When the agents identify issues such as inconsistent problem interpretations, malformed mathematical structures, or code generation errors, they automatically route the workflow back to the responsible agent
with specific feedback. This iterative refinement mechanism ensures that the system self-corrects common modeling mistakes while maintaining full transparency: users can inspect each iteration and understand exactly what was corrected and why. Unlike prior works which output solely an optimal value, our system allows the user to validate each step of the modeling process.

\subsection{Related Works}

Due to the high complexity of the modeling task, an active research area has started with \citep{NL4Opt}. Existing work can be roughly divided in two main categories: multi-agent frameworks and learning-based approaches. 

%The research in the multi-agent field is mainly motivated by proposing automated systems with structured subtasks for each agent. Common directions proposes  different tools or agents that mimic the human modeling process or introducing specialized OR domain.
Chain-of-Experts (CoE) \citep{chainexp} proposes a multi-agent framework where each of five agents is assigned to a specific role, orchestrating, interpreting, modeling, programming and evaluating. OptiMUS \citep{optimus} also proposes a structured list of LLM agents to model and solve optimization problems with a predefined workflow and tailored prompts. Natural-language to Executable Mathematical Optimization (NEMO) \citep{nemo} proposes a multi-agent architecture using Autonomous Coding Agents (ACAs) that operate within sandboxed environments, which introduces potential latency and infrastructure overhead. %On the other hand, all code produced by NEMO is executable by construction, enabling automated validation and repair without manual intervention. 
In order to eliminate redundancy of these autonomous agents, the paper ORThought \citep{orthought} proposes a dual-agent framework incorporating expert-level modeling principles via chain-of-thought reasoning. %It performs single-pass formulation generation and does not refine the mathematical model once it is produced. 

The learning-based direction makes use of the powerful capability of foundational models and specialize them to the optimization modeling task. LLMOPT \citep{llmopt} proposes a multi-instruction supervised fine-tuning, along with model alignment to further enhance accuracy and reduce the risk of hallucinations. ORLM \citep{orlm} is a training framework specifically designed to use open-source LLMs for automated optimization modeling. %It provides a reproducible pipeline for training and data generation during the instruction tuning phase that practitioners can adapt to their own organizational contexts.

As noted in the introduction, explanation is essential in OR, especially in high-stakes settings where decisions have significant operational or societal impact. %The modeling process should therefore be transparent and trustworthy, so stakeholders can understand, validate, and challenge model assumptions and outputs. 
Despite growing interest, research progress in this direction remains limited. Recent contributions include \citep{li2023largelanguagemodelssupply}, which supports what-if analysis in supply chain optimization, and \citep{ERWIG2024101272}, which explains solutions by contrasting them with plausible alternatives. More recently, Explainable Operations Research (EOR) was introduced in \citep{ICLR2025_a48e5877} as a bipartite-graph-based framework to quantify and communicate the effects of model changes.

\section{The architecture}
\begin{figure}[!t]
\centering
\includegraphics[width=1.0\textwidth]{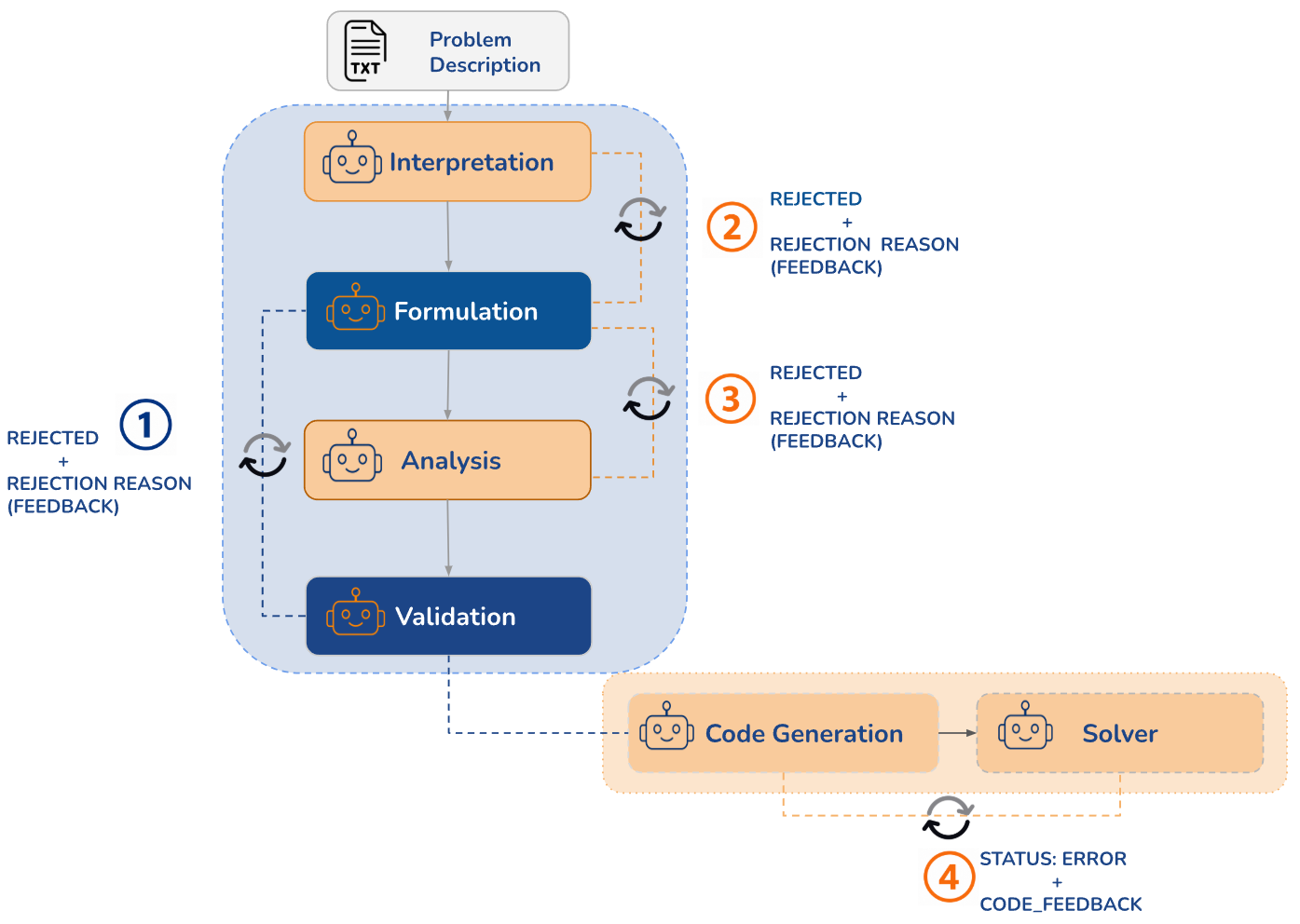}
\caption{OptiAgent: a multi-agent framework with internal feedback loops for iterative refinement of the optimization formulation and generated code.} \label{fig1}
\end{figure}

Our framework consists of six specialized agents organized in a sequential pipeline with targeted feedback loops as showed in Figure~\ref{fig1}. This design emerged from analyzing the distinct cognitive tasks required in
optimization modeling: understanding the problem domain, formalizing mathematical relationships, validating structural correctness, generating executable code, and computing solutions. Unlike existing multi-agent systems that rely on generic roles or homogeneous ensembles, each agent has a specific responsibility aligned with the stages of expert human modelers' workflow.

The six agents operate on a shared state (AgentState) that accumulates intermediate outputs, feedback signals, and iteration counters as the workflow progresses. The pipeline follows a dependency chain interpretation that provides the entities needed for variable definition, formulation enables complexity assessment, analysis informs validation criteria, validation prevents implementing invalid models, and syntactically correct code is required before solving.

To handle errors without human intervention, the framework implements four targeted feedback loops that route the workflow back to specific upstream agents upon error detection. If the validation agent detects mathematical errors such as non-linearity in LP formulations or constraint contradictions it  routes back to formulation (Loop 1). When the analysis agent
identifies structural issues such as undefined variables or missing constraints, it triggers a loop back to the formulation agent (Loop 3).

More fundamental misunderstandings of the problem cause formulation to route all the way back to the interpretation agent (Loop 2). Finally, when the solver encounters syntax errors or runtime exceptions during code execution, it triggers a loop back to the code generation agent (Loop 4). All loops are constrained to a maximum of three
iterations, ensuring the system self-corrects common errors while avoiding infinite refinement cycles.

%\red{Não acho que a definição de Loop2, Loop3, etc estejam claras nessa parte do texto. Refrasear: it triggers a loop back to the formulation agent, which we call Loop 2... Aliás, cadê o Loop 1? Tudo isso ficaria muito mais claro com uma figura.}

%\subsection{Agent Specifications / Agents Design}

The Interpretation agent converts the natural language problem description into a structured representation, capturing the optimization intent, objectives, entities, and constraints in plain language without premature formalization. The Formulation agent then turns this into a complete mathematical model in the five-element formulation. Next, the Analysis agent characterizes the problem, e.g. as LP or MILP, based on its complexity and structure given by variable types, constraints, and objective. This guides validation and solver selection.

The Validation Agent acts as a quality gate at the syntactic (all variables defined), semantic (constraints match intent), and mathematical (feasibility and boundedness) levels, returning \texttt{APPROVED}/\texttt{REJECTED} with feedback that names the responsible agent (e.g., "FORMULATION: missing capacity constraints"). Once approved, the Code Generation Agent translates the formulation into executable Python code, selecting a library (Pyomo, PuLP, or OR-Tools) by problem type. Finally, the Solver Agent runs this code in a sandboxed environment and parses the output to report solution status, objective value, and variable values, distinguishing code errors (syntax, runtime) to route feedback to the code generation agent. In  Appendix~\ref{sec:prompttemp}, we present the prompt templates for the Formulation and Validation agents, which served as the basis for the design of all remaining agent prompts.

\paragraph{Five-element formulation of optimization problems.}

Keeping in mind that our framework prioritizes the mathematical modeling step, we adopt the universal representation from \citep{llmopt}, which structures a NL description of an optimization problem into key five-elements: \textit{Sets, Parameters, Variables, Objective} and \textit{Constraints} to better support LLM reasoning. 
The \textit{Objective} represents the goal, that is, what one wants to accomplish by solving that optimization problem. \textit{Variables} denotes the unknown that one is trying to solve the problem for. \textit{Parameters} are the specific values given by the problem description. \textit{Constraints} is a vector-valued function that contains what kind of restrictions the decision variable must satisfy. Finally, \textit{Sets} provide detailed information of indices and descriptions. This decomposition  promotes a more accurate model while preserving essential mathematical descriptions. 

% Formally, an optimization problem can be abstracted into mathematical structures by the following expression
% \begin{equation}\label{eq:minprob}
%     \min_{x\in\mathcal{X}} f(x) \text{ s.t. } G(x)\leq x
% \end{equation}
% where $f:\mathcal{X}\subset\R^d\to\R$ represents the objective function, $x\in\R^d$ is the decision variable, $\mathcal{X}\subset\R^d$ denotes the feasible domain and $G:\R^d\to\R^m$ is the constraint function.

% In this work we adopt an universal representation of \eqref{eq:minprob} proposed by \citep{llmopt}. The goal is to translate \eqref{eq:minprob} into key elements that facilitate the LLM reasoning step. This formulation consists in extracting from the NL description the following five-elements: \textit{Sets, Parameters, Variables, Objective} and \textit{Constraints}. 

%  this five-element formulation
%\red{fico com a sensação que isso deveria estar lá na seção 2. 5-elements ficaram faltando lá atrás para entender melhor o que cada um dos agentes fazia. Isso é basicamente setup do problema, que já precisa estar claro para o leitor quando vamos falar dos agentes.}

\section{Experiments}

\subsection{Experimental Setup}

For the sake of comparison, we produced a baseline experiment to reproduce a vanilla usage of LLMs by a non-expert user attempting to solve optimization problems. The pipeline followed four main steps: (1) a first API call to extract a structured five-element model representation from the natural language problem description; (2) a second API call to generate the corresponding solver code from that representation; (3) a subprocess execution to run the generated code and collect the solver output log; and (4) a final API call to extract the optimal value from the output log given by the solver. If the generated code fails to execute, the pipeline retries code generation up to three times before moving on. The model generates code using the Pyomo optimization library and selects the appropriate solver based on the problem type inferred from the NL description. 

To rigorously evaluate the impact of structured guidance, we consider two baseline settings. First, no dedicated system prompt is provided, allowing the model to rely solely on its pre-trained knowledge and implicit reasoning to infer the five-element formulation. This setup serves to assess the model’s out-of-the-box capability to translate natural-language optimization problems into formal representations without external scaffolding. In contrast, we introduce a guided setting in which a dedicated system prompt explicitly explains the five-element formulation, steering the model toward a more consistent and structured extraction of the representation from each problem description. %This comparison enables us to isolate the contribution of explicit prompting and determine whether structured guidance meaningfully improves robustness and accuracy.
%Furthermore, we added a dedicated system prompt that explains the five-element formulation, guiding the model to consistently extract the structured five-element representation from each natural-language problem description.

We follow the experimental protocols of prior work used for benchmarking. Prior studies are inconsistent in their reporting: some average results over multiple runs, while others report single-run outcomes, and the number of runs differs across works. To preserve comparability with these baselines, we present the headline results in Table~1 without including variability measures.

\subsubsection{Baseline models.} We evaluated state-of-the-art models known for their strong performance reasoning, coding and mathematical tasks: GPT-5.2 and GPT-5.4 \citep{gpt5}, 
Claude Sonnet 4.5 \citep{claude-sonnet-45}, and DeepSeek-R1 \citep{deepseek-r1}. To ensure consistency across experiments we fixed GPT-5.2 as the answer extractor in all three cases, regardless of which model was used as the modeler. All models were queried with temperature set to 0 to produce more deterministic outputs across runs, which is beneficial for knowledge-intensive problems as observed in \citep{chainexp,orthought}.
%We evaluated our multi-agent optimization system on 4 benchmark dataset, which comprises optimizations problems including diverse domains such aircraft assignment, production scheduling, resource allocation, and network optimization. Each problem in the dataset includes a natural language description with numerical data and the expected optimal solution. The system was configured with Claude Sonnet 4.5 as the underlying LLM, operating through a six-agent pipeline consisting of interpretation, formulation, analysis, validation, code generation, and solver components. Notably, we implemented a simplified solver architecture that eliminates LLM inference at the final execution stage parsing only the solver output.

\subsubsection{OptiAgent.}
Our framework was implemented using LangGraph, which supports the construction of stateful, multi-agent workflows executed as directed acyclic graphs (DAGs). To isolate the effects of the multi-agent architecture and the proposed feedback mechanisms, all experiments were conducted using a fixed LLM backbone, Claude Sonnet 4.5 and GPT 5.4 with temperature set to 0.

\subsection{Datasets}

We conducted experiments on four real-world optimization datasets widely used in the literature, ComplexOR, IndustryOR, LogiOR and NLP4LP. We used the corrected and re-annotated versions\footnote{Available at \url{https://github.com/ZJU-TSELab/ORThought}.} curated by \citep{orthought}. Comprising a total of 455 problems including Linear Programming (LP), Integer Linear Programming (ILP), Mixed-Integer Linear Programming (MILP),
and Nonlinear Programming (NLP) these four datasets are described in Table~\ref{tab1}.

\paragraph{\textbf{ComplexOR.}}
The ComplexOR dataset proposed by \citep{chainexp} was created by three OR specialists. It covers a wide range of domains, such as scheduling, assignment and revenue maximization.

\paragraph{\textbf{IndustryOR.}}
The IndustryOR dataset \citep{orlm} was specifically designed for automated optimization modeling. It contains data derived from real-world cases focusing in practical applications. It covers diverse scenarios from different industries, including education,
transportation, and finance.

\paragraph{\textbf{LogiOR.}} One of the most recent datasets introduced for the automated optimization modeling task.
A benchmark curated by \citep{orthought} comprising logistics and supply chain optimization problems. It was developed under the guidance of three Operations Research (OR) experts. 

\paragraph{\textbf{NLP4P.}}
The NLP4LP dataset \citep{optimus} consists of optimization problems sourced from optimization textbooks and lecture notes covering areas such as facility location,
network flow, scheduling, and portfolio management. 

% \begin{table}[h]
% \centering
% \caption{Problem types per dataset.}
% \label{tab1}
% \begin{tabular}{|l|c|c|c|c|}
% \hline
% \textbf{Problem type} & \textbf{ ComplexOR } & \textbf{ IndustryOR } & \textbf{ LogiOR } & \textbf{ NLP4LP } \\
% \hline
% LP   & 10 & 18 & 22 & 55 \\
% ILP  & 7 & 47 & 43 & 206 \\
% MILP & 1 & 14 & 11 & 1 \\
% NLP  & 0 & 4 & 16 & 0 \\
% Total & 18 & 83 & 92 & 262\\
% \hline
% \end{tabular}
% \end{table}

\begin{table}[h]
\centering
\setlength{\tabcolsep}{8pt}
\caption{Problem types per dataset.}
\label{tab1}
\begin{tabular}{lcccc}
\toprule
& \textbf{ComplexOR} & \textbf{IndustryOR} & \textbf{LogiOR} & \textbf{NLP4LP} \\
\midrule
LP    & 10 & 18 & 22 & 55 \\
ILP   & 7  & 47 & 43 & 206 \\
MILP  & 1  & 14 & 11 & 1 \\
NLP   & 0  & 4  & 16 & 0 \\
\midrule
Total & 18 & 83 & 92 & 262 \\
\bottomrule
\end{tabular}
\end{table}

%industryOR gpt 5.4 (70/83)
%complexOR gpt 5.4 (18/18)
%NLP4LP gpt 5.4 (00/262)
%NLP4LP gpt 5.4 (00/92)

\subsection{Results}

We use solving accuracy (SA) as our evaluation metric, which measures whether the generated optimization model produces solutions that match ground truth objective values. Table~\ref{tab:overall_results} reports SA across multiple methods, including both baseline (w/o system prompt) and variants.

SA values for ORLM, Coe and ORThought are taken from \citep{orthought}. For LLMOPT, we report values from the original paper \citep{llmopt} on the NLP4LP, IndustryOR, and ComplexOR datasets, and run the corresponding model\footnote{Available at \url{https://huggingface.co/ant-opt/LLMOPT-Qwen2.5-14B}} on LogiOR. We also report results for NEMO directly taken from \citep{nemo} on the NLP4LP, IndustryOR, and ComplexOR datasets.
%\red{me fiz várias perguntas sobre o resultado (por exemplo: qual modelo foi usado no OptiAgent?) Só pra descobrir que isso é dito depois. Pra mim, toda a descrição do sistema tem que vir antes dos resultados}

\begin{table}
\caption{SA across different methods on 4 datasets. Bold: best performance.}
    \centering
    \setlength{\tabcolsep}{4pt}
    \begin{tabularx}{\textwidth}{XXXXX}
        \toprule
        %\textbf{Method} 
        & \textbf{NLP4LP} & \textbf{IndustryOR} & \textbf{LogiOR } & \textbf{ComplexOR} \\
        \midrule
        GPT 5.2 w/o & 74.24\% & 66.27\% & 42.39\% & 66.67\% \\
        GPT 5.4 w/o & 80.46\% & 67.07\% & 60.87\% & 83.33\% \\
        Sonnet w/o & 78.41\% & 63.86\% & 44.57\% & 61.11\%\\
        Deepseek w/o & 64.02\% & 46.99\% & 19.57\% & 72.22\%\\
        %\midrule
        \midrule
        GPT 5.2 & 83.46\% & 72.297\% & 52.17\% & 77.78\% \\
        GPT 5.4 & 81.61\% & 65.85\% & 54.35\% & 83.33\% \\
        Sonnet 4.5 & 81.82\% & 67.47\% & 51.09\% & 55.56\%\\
        Deepseek-R1 & 76.89\% & 51.81\% & 26.09\% & 50.00\%\\
        \midrule
        ORLM & 67.8\% & 32.13\% & 15.58 \% & 77.78\% \\
        LLMOPT & 83.8\% & 46\% & 18.48 \% & 72.7\% \\
        ORThought & \textbf{89.02} \%& 57.83\% & 46.01\% & 77.78\% \\
        NEMO & 81.4 \%& 63.0\% & - & 77.8\% \\
        CoE &75\%& 40.96\%&34.78\%&55.56\%\\
        OptiAgent (Sonnet 4.5) & 76.14\% & 71.08\% & 68.48\% & \textbf{100}\% \\
        OptiAgent (GPT 5.4) & 87.12\% & \textbf{84.34}\% & \textbf{70.65}\% & \textbf{100}\% \\
        \bottomrule
    \end{tabularx}
    
    \label{tab:overall_results}
\end{table}

Overall, GPT-5.4 consistently achieves the strongest performance among the base models, reaching 83.33\% on ComplexOR and 80.46\% on NLP4LP in the baseline setting. However, the impact of the system prompt is not uniformly positive. While GPT-5.2 benefits from prompting across all datasets (e.g., +9.1 points on ComplexOR and +10 points on LogiOR), GPT-5.4 exhibits mixed behavior, with slight improvements on NLP4LP (+1.15) but performance drops on IndustryOR and LogiOR. Similarly, Sonnet and DeepSeek show inconsistent or negative gains when the structured prompt is introduced, particularly on ComplexOR, where performance decreases by more than 5 points for Sonnet and over 20 points for DeepSeek.

This behavior becomes more pronounced when analyzing differences across datasets. LogiOR is clearly the most challenging benchmark, combining longer and more complex natural-language descriptions with tighter and more intricate constraints. In addition, LogiOR is the most recent dataset, making it less likely that similar problem instances were seen during pretraining. This setting therefore better captures out-of-distribution generalization, where models must rely on reasoning rather than memorization. Notably, this is precisely where OptiAgent shows the largest gains, substantially outperforming all base models. In contrast, ComplexOR represents the simplest setting, with fewer instances and comparatively easier formulations, which is reflected in the overall higher and less variable performance across methods. More broadly, the older datasets (NLP4LP and ComplexOR) consistently yield the highest scores, suggesting that temporal familiarity and dataset overlap may play a role in model performance.

Across all datasets, OptiAgent emerges as the most robust and consistently high-performing approach, outperforming both base LLM configurations and prior OR-specific methods in the majority of settings. In particular, its gains are most pronounced on LogiOR and IndustryOR, where it substantially exceeds all baselines, highlighting its ability to handle more complex, recent, and previously unseen problem formulations. Furthermore, OptiAgent achieves perfect accuracy on ComplexOR, demonstrating that its multi-stage design does not sacrifice performance even in simpler settings. 

\subsection{Additional Analysis}

\subsubsection{Execution Time.}
Our system implements autonomous feedback loops for self-correction. To prevent infinite execution, we
enforce threshold limits of 3 max iterations per correction cycle. Execution time is measured immediately before the first LLM invocation and ends after token extraction from the final agent response, encompassing all feedback loop iterations but excluding solver runtime (where the problem complexity plays a big role).

\begin{table}[t]
\caption{Runtime efficiency and feedback loop analysis of OptiAgent.}
\centering
\setlength{\tabcolsep}{3pt}
\begin{tabularx}{\textwidth}{lcccc}
\toprule
  & \textbf{Avg Time (s)} & \textbf{\#Loops} & \textbf{\#Problems w/ Loops} & \textbf{Loop Rate} \\
\midrule
NLP4LP     & 43.85  & 96  & 87  & 33.1\% \\
IndustryOR  & 95.34  & 93  & 51  & 61.4\% \\
ComplexOR   & 53.90  & 9   & 5   & 27.8\% \\
LogiOR     & 122.60 & 105 & 49  & 55.1\% \\
\bottomrule
\end{tabularx}
\label{tab:runtime_feedback}
\end{table}

In table~\ref{tab:runtime_feedback} we report runtime efficiency and feedback loop analysis of OptiAgent using Claude Sonnet 4.5 as backbone. Our framework processed all NLP4LP  263 problems in 3.20 hours achieving a rate of 87 of 263 loops per problem. The IndustryOR dataset showed intermediate efficiency with total runtime of 2.20 hours. It presented feedback loops in 61\% of cases. Despite the name "Complex", ComplexOR solved all of its 18 problems in 16.2 minutes triggering only 9 feedback loops across 5 problems, the lowest proportion among all datasets, indicating that our framework understood and formulated these problems more directly. Finally for the LogiOR dataset, notably the one with greater complexity, our framework took 3.03 hours of total runtime. With 105 feedback loops triggered across 49 problems, it demonstrates the need for more corrections.

\subsubsection{Feedback Loops.}
\begin{figure}[t]
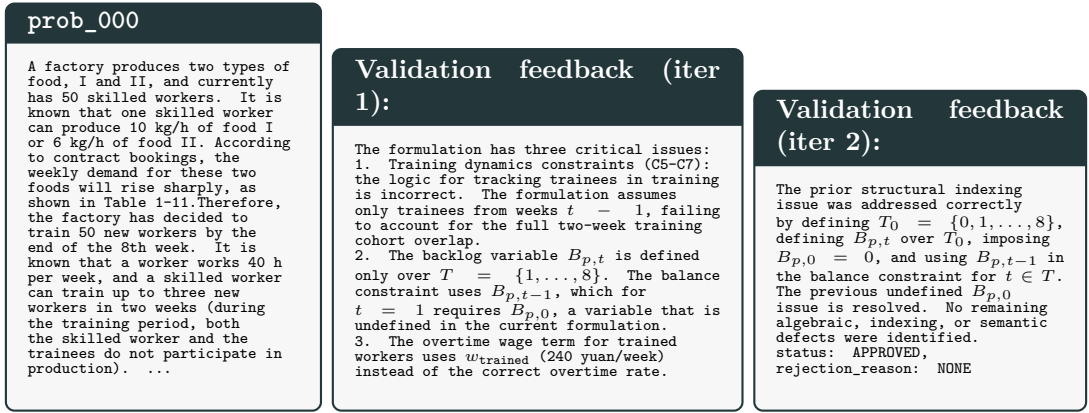

    \centering
    \begin{subfigure}[b]{0.285\textwidth}
        \begin{bluebox}[\texttt{prob\_000}]
        \begin{spacing}{0.81} 
        \ttfamily\tiny\linespread{0.9}
        A factory produces two types of food, I and II, and currently has 50 skilled workers. It is known that one skilled worker can produce 10 kg/h
        of food I or 6 kg/h of food II. According to contract bookings, the weekly demand for these two foods will rise sharply, as shown in Table 1-11.Therefore, the factory has decided to train 50 new workers by the end of the 8th week. It is known that a worker works 40 h per week, and a skilled worker can train up to three new workers in two weeks (during the training period, both the skilled worker and the trainees do not participate in production). ...
        \end{spacing}
    \end{bluebox}
    \end{subfigure}
    \begin{subfigure}[b]{0.37\textwidth}
    \begin{bluebox}[Validation feedback (iter 1):]
    \begin{spacing}{0.82} 
    \ttfamily\tiny\linespread{0.9}
    The formulation has three critical issues:\\
    1. Training dynamics constraints (C5--C7): the logic for tracking trainees in training is incorrect. The formulation assumes only trainees from weeks $t-1$, failing to account for the full two-week training cohort overlap.\\
    2. The backlog variable $B_{p,t}$ is defined only over $T = \{1,\dots,8\}$. The balance constraint uses $B_{p,t-1}$, which for $t=1$ requires $B_{p,0}$, a variable that is undefined in the current formulation. \\
    3. The overtime wage term for trained workers uses $w_{\text{trained}}$ (240 yuan/week) instead of the correct overtime rate.
         \end{spacing}
    \end{bluebox}
    \end{subfigure}
    \begin{subfigure}[b]{0.3\textwidth} 
       \begin{bluebox}[Validation feedback (iter 2):]
       \begin{spacing}{0.82}  
        \ttfamily\tiny%\linespread{0.7}
        The prior structural indexing issue was addressed correctly by defining $T_0 = \{0,1,\dots,8\}$, defining $B_{p,t}$ over $T_0$, imposing $B_{p,0}=0$, and using $B_{p,t-1}$ in the balance constraint for $t \in T$. The previous undefined $B_{p,0}$ issue is resolved. No remaining algebraic, indexing, or semantic defects were identified.\\
        \texttt{status: APPROVED}, \texttt{rejection\_reason: NONE}
         \end{spacing}
    \end{bluebox}
    \end{subfigure}
    %\vspace{-4pt}
    \caption{An instance from IndustryOR triggered loops during validation stage. Feedbacks returned by validation agent.
    }
    \label{fig:prob000industryOR}
\end{figure}

An example where a loop was triggered during the validation stage is the instance \texttt{prob\_000} taken from the IndustryOR dataset, as illustrated in Figure~\ref{fig:prob000industryOR}.

On Loop 1, after receiving the mathematical formulation of the problem, the Validation agent identified three mathematical errors and rejected the result, triggering a return to the Formulation agent.

Without the loop, the generated code would have referenced the undefined variable \(B_{p,0}\), potentially causing a runtime error or producing a mathematically incorrect model. The feedback mechanism caught this issue during the \emph{formulation} stage, \emph{before} any code generation took place. After revision, the Validation agent returned \texttt{APPROVED}, meaning that the validator did not identify any remaining errors.

%\red{Dado que, nas contribuições iniciais, foi dada tanta importância à interpretabilidade, acho um ponto muito fraco do paper não ter nenhum exemplo, métrica ou seja o que for que valide o fato do framework multi-agente ter levado a maior interpretabilidade. De modo geral, achei a introdução bem desconexa com o que está sendo apresentado em termos de resultados. Se queremos afirmar acurácia estado da arte, isso já deveria estar lá no início. Mas esse nem é o caso, não estamos comparando com o Nemo, que me parece que performa melhor em vários casos aqui.}

\section{Discussion}
%\red{limitacoes, uso de outros modelos nos agentes, falar do problema dos dados, adicionar feedback humano de problemas escritos por pessoas normais}

%\red{falar q n repetimos os experimentos entao n reportamos media das metricas}
Our approach achieves the best results on three of the four datasets, which we attribute to error recovery, task specialization, and iterative refinement enabled by the multi-agent architecture. Unlike single-pass baselines, our framework can identify and correct intermediate mistakes, leading to more robust performance.

While prompted LLM baselines perform well, their pipelines remain largely opaque, making it difficult to diagnose whether errors arise from extraction, code generation, or parsing. In contrast, our approach provides interpretable intermediate representations, traceable feedback loops, and inspectable artifacts, improving transparency and reliability.
Notably, these gains occur even though explicit structured prompting alone does not consistently improve performance, indicating that architectural decomposition, not prompt engineering, is the key driver of improvements. 

Finally, experiments were conducted with a single run per configuration, and thus do not capture output variability. However, performance trends and relative rankings remain consistent across datasets, suggesting that the results are robust. Future work should incorporate repeated trials to assess variance more rigorously.

\section{Conclusion}

In this paper, we presented OptiAgent, a multi-agent framework designed for end-to-end optimization. Our approach achieves the best results on three of the four datasets and remains highly competitive on the remaining benchmarks. We hypothesize that these gains stem from three architectural properties: (1) error detection and recovery, where validation and analysis agents identify formulation issues that may go unnoticed in single-shot generation; (2) specialized reasoning, since each agent targets a well-scoped subtask (interpretation, formulation, validation) rather than performing end-to-end generation in one step; and (3) iterative refinement, in which feedback loops support incremental corrections instead of requiring perfect first-pass outputs. 
By focusing on the modeling phase of the optimization pipeline, our architecture effectively leverages the reasoning capabilities of LLMs. Additionally, the inter-agent verification loops promote transparency and enhance the trustworthiness of the overall process. Overall, these results suggest that multi-agent coordination with feedback can provide substantial benefits over single-shot approaches, particularly on problems with more complex constraint structures and objective specifications.

\paragraph{Future work.}
We aim to analyze the contribution of each component of the proposed framework through ablation studies. In particular, we plan to remove or modify individual agents to quantify their impact on performance. Additionally, we will ablate the feedback loop mechanism to better understand the role of iterative refinement and determine whether improvements stem primarily from task decomposition or from the ability to correct intermediate errors.

Another key direction is the development of interpretability metrics tailored to OR modeling tasks, including measures of formulation clarity and alignment between natural-language descriptions and mathematical models. We also aim to incorporate human evaluations with experts and non-expert users to assess how well the system supports understanding and trust. Finally, future experiments will include multi-run evaluations to account for the stochasticity of LLM outputs and provide a more robust assessment of performance and generalization.
%\begin{credits}
\paragraph{Disclosure of Interests} Any opinions, findings, conclusions or recommendations expressed in this material are those of the authors and do not necessarily reflect the views of Itaú Unibanco and Instituto de Ciência e Tecnologia Itaú. This document is not and does not constitute or intend to constitute investment advice or any investment service. It is not and should not be deemed to be an offer to purchase or sell, or a solicitation of an offer to purchase or sell, or a recommendation to purchase or sell any securities or other financial instruments. In addition, all data used in this study comply with the Brazilian General Data Protection Law.
\bibliographystyle{plainnat}
\bibliography{references}

@article{ERWIG2024101272,
title = {Explanations for combinatorial optimization problems},
journal = {Journal of Computer Languages},
volume = {79},
pages = {101272},
year = {2024},
issn = {2590-1184},
doi = {https://doi.org/10.1016/j.cola.2024.101272},
url = {https://www.sciencedirect.com/science/article/pii/S2590118424000157},
author = {Martin Erwig and Prashant Kumar}
}

@misc{li2023largelanguagemodelssupply,
      title={Large Language Models for Supply Chain Optimization}, 
      author={Beibin Li and Konstantina Mellou and Bo Zhang and Jeevan Pathuri and Ishai Menache},
      year={2023},
      eprint={2307.03875},
      archivePrefix={arXiv},
      primaryClass={cs.AI},
      url={https://arxiv.org/abs/2307.03875}, 
}

@article{DEBOCK2024249,
title = {Explainable AI for Operational Research: A defining framework, methods, applications, and a research agenda},
journal = {European Journal of Operational Research},
volume = {317},
number = {2},
pages = {249-272},
year = {2024},
issn = {0377-2217},
doi = {https://doi.org/10.1016/j.ejor.2023.09.026},
url = {https://www.sciencedirect.com/science/article/pii/S0377221723007294},
author = {Koen W. {De Bock} and Kristof Coussement and Arno De Caigny and Roman Słowiński and Bart Baesens and Robert N. Boute and Tsan-Ming Choi and Dursun Delen and Mathias Kraus and Stefan Lessmann and Sebastián Maldonado and David Martens and María Óskarsdóttir and Carla Vairetti and Wouter Verbeke and Richard Weber}
}

@inproceedings{ICLR2025_a48e5877,
 author = {Zhang, Yansen and Kang, Qingcan and YU, Wing Yin and HaileiGong and Fu, Xiaojin and Han, Xiongwei and Zhong, Tao and Ma, Chen},
 booktitle = {International Conference on Learning Representations},
 editor = {Y. Yue and A. Garg and N. Peng and F. Sha and R. Yu},
 pages = {65698--65722},
 title = {Decision Information Meets Large Language Models: The Future of Explainable Operations Research},
 url = {https://proceedings.iclr.cc/paper_files/paper/2025/file/a48e5877c7bf86a513950ab23b360498-Paper-Conference.pdf},
 volume = {2025},
 year = {2025}
}

@inproceedings{chainexp,
 author = {Xiao, Ziyang and Zhang, Dongxiang and Wu, Yangjun and Xu, Lilin and Wang, Yuan and Han, Xiongwei and Fu, Xiaojin and Zhong, Tao and Zeng, Jia and Song, Mingli and Chen, Gang},
 booktitle = {International Conference on Learning Representations},
 editor = {B. Kim and Y. Yue and S. Chaudhuri and K. Fragkiadaki and M. Khan and Y. Sun},
 pages = {48519--48537},
 title = {Chain-of-Experts: When LLMs Meet Complex Operations Research Problems},
 url = {https://proceedings.iclr.cc/paper_files/paper/2024/file/d45ee77826332c100a1e15f7765b99ff-Paper-Conference.pdf},
 volume = {2024},
 year = {2024}
}

@InProceedings{optimus,
  title = 	 {{O}pti{MUS}: Scalable Optimization Modeling with ({MI}){LP} Solvers and Large Language Models},
  author =       {Ahmaditeshnizi, Ali and Gao, Wenzhi and Udell, Madeleine},
  booktitle = 	 {Proceedings of the 41st International Conference on Machine Learning},
  pages = 	 {577--596},
  year = 	 {2024},
  editor = 	 {Salakhutdinov, Ruslan and Kolter, Zico and Heller, Katherine and Weller, Adrian and Oliver, Nuria and Scarlett, Jonathan and Berkenkamp, Felix},
  volume = 	 {235},
  series = 	 {Proceedings of Machine Learning Research},
  month = 	 {21--27 Jul},
  publisher =    {PMLR},
  url = 	 {https://proceedings.mlr.press/v235/ahmaditeshnizi24a.html},
  
}

@InProceedings{NL4Opt,
  title = 	 {NL4Opt Competition: Formulating Optimization Problems Based on Their Natural Language Descriptions},
  author =       {Ramamonjison, Rindranirina and Yu, Timothy and Li, Raymond and Li, Haley and Carenini, Giuseppe and Ghaddar, Bissan and He, Shiqi and Mostajabdaveh, Mahdi and Banitalebi-Dehkordi, Amin and Zhou, Zirui and Zhang, Yong},
  booktitle = 	 {Proceedings of the NeurIPS 2022 Competitions Track},
  pages = 	 {189--203},
  year = 	 {2022},
  editor = 	 {Ciccone, Marco and Stolovitzky, Gustavo and Albrecht, Jacob},
  volume = 	 {220},
  series = 	 {Proceedings of Machine Learning Research},
  month = 	 {28 Nov--09 Dec},
  publisher =    {PMLR},
  url = 	 {https://proceedings.mlr.press/v220/ramamonjison23a.html},
}

@article{orlm,
   title={ORLM: A Customizable Framework in Training Large Models for Automated Optimization Modeling},
   volume={73},
   ISSN={1526-5463},
   url={http://dx.doi.org/10.1287/opre.2024.1233},
   DOI={10.1287/opre.2024.1233},
   number={6},
   journal={Operations Research},
   publisher={Institute for Operations Research and the Management Sciences (INFORMS)},
   author={Huang, Chenyu and Tang, Zhengyang and Hu, Shixi and Jiang, Ruoqing and Zheng, Xin and Ge, Dongdong and Wang, Benyou and Wang, Zizhuo},
   year={2025},
   month=nov, pages={2986–3009} }

@article{nemo,
  title={NEMO: Execution-Aware Optimization Modeling via Autonomous Coding Agents},
  author={Song, Yang and Vyas, Anoushka and Wei, Zirui and Pakazad, Sina Khoshfetrat and Ohlsson, Henrik and Neubig, Graham},
  journal={arXiv preprint arXiv:2601.21372},
  year={2026}
}

@inproceedings{
llmopt,
title={{LLMOPT}: Learning to Define and Solve General Optimization Problems from Scratch},
author={Caigao JIANG and Xiang Shu and Hong Qian and Xingyu Lu and JUN ZHOU and Aimin Zhou and Yang Yu},
booktitle={The Thirteenth International Conference on Learning Representations},
year={2025},
url={https://openreview.net/forum?id=9OMvtboTJg}
}

@inproceedings{ijcai2025p1192,
  title     = {A Survey of Optimization Modeling Meets LLMs: Progress and Future Directions},
  author    = {Xiao, Ziyang and Xie, Jingrong and Xu, Lilin and Guan, Shisi and Zhu, Jingyan and Han, Xiongwei and Fu, Xiaojin and Yu, WingYin and Wu, Han and Shi, Wei and Kang, Qingcan and Duan, Jiahui and Zhong, Tao and Yuan, Mingxuan and Zeng, Jia and Wang, Yuan and Chen, Gang and Zhang, Dongxiang},
  booktitle = {Proceedings of the Thirty-Fourth International Joint Conference on
               Artificial Intelligence, {IJCAI-25}},
  publisher = {International Joint Conferences on Artificial Intelligence Organization},
  editor    = {James Kwok},
  pages     = {10742--10750},
  year      = {2025},
  month     = {8},
  note      = {Survey Track},
  doi       = {10.24963/ijcai.2025/1192},
  url       = {https://doi.org/10.24963/ijcai.2025/1192},
}

@misc{orthought,
      title={ORThought: Benchmarking and Automating Logistics Optimization Modeling}, 
      author={Beinuo Yang and Qishen Zhou and Junyi Li and Chenxing Su and Panagiotis Angeloudis and Simon Hu},
      year={2026},
      eprint={2508.14410},
      archivePrefix={arXiv},
      primaryClass={cs.AI},
      url={https://arxiv.org/abs/2508.14410}, 
}

@article{deepseek-r1,
  title   = {DeepSeek-R1: Incentivizing Reasoning Capability in LLMs via Reinforcement Learning},
  author  = {DeepSeek-AI},
  journal = {arXiv preprint arXiv:2501.12948},
  year    = {2025},
  url     = {https://arxiv.org/abs/2501.12948}
}

@techreport{claude-sonnet-45,
  title       = {Claude Sonnet 4.5},
  author      = {Anthropic},
  institution = {Anthropic},
  year        = {2025},
  url         = {https://www.anthropic.com}
}

@misc{gpt5,
      title={OpenAI GPT-5 System Card}, 
      author={OpenAI},
      year={2025},
      eprint={2601.03267},
      archivePrefix={arXiv},
      primaryClass={cs.CL},
      url={https://arxiv.org/abs/2601.03267}, 
}
\newpage
\appendix
\section{Prompt Templates}\label{sec:prompttemp}
%In this section, we present the prompt templates for the Formulation and Validation agents, which served as the basis for the design of all remaining agent prompts.

\definecolor{promptblue}{RGB}{59,89,182}
\lstdefinestyle{pythonprompt}{
 language=Python,
 basicstyle=\footnotesize\ttfamily,
 breaklines=true,
 breakatwhitespace=true,
 columns=fullflexible,
 keepspaces=true,
 showstringspaces=false
}
%%%%%%%
%%%% FORMULATION 
\begin{tcblisting}{
  enhanced,
  colback=white,
  colframe=promptblue!70,
  boxrule=0.8pt,
  sharp corners,
  width=\dimexpr\textwidth+2cm\relax,
  enlarge left by=-1.5cm,
  enlarge right by=-1.5cm,
  left=3mm,
  right=3mm,
  top=3mm,
  bottom=3mm,
  boxsep=1mm,
  listing only,
  listing options={
    basicstyle=\footnotesize\ttfamily,
    breaklines=true,
    breakatwhitespace=false,
    columns=flexible,
    keepspaces=true,
    showstringspaces=false,
    breakindent=0pt,
    postbreak={},
    xleftmargin=0pt,
    xrightmargin=0pt,
    resetmargins=true
  },
  overlay={
    \node[
      fill=promptblue,
      text=white,
      font=\bfseries\scriptsize\ttfamily,
      rounded corners=1mm,
      anchor=west
    ] at ([xshift=10mm,yshift=-2mm]frame.north west)
    {[SYSTEM PROMPT]};
  }
}
You are a Mathematical Formulation Agent specializing in Operations Research.

Input
-----
You receive a structured interpretation of an optimization problem (the output of the Interpretation Agent).

Task
----
Formulate the optimization problem following the standard 5-element structure.

Step 1: Build the 5-element formulation
---------------------------------------
1. Sets: Index sets (e.g., I = set of customers, T = time periods);
2. Parameters: Known data/constants with numeric values and domain/indices (e.g., c_ij = cost, d_i = demand).
3. Decision Variables: Variables to be decided, with domain specification (Binary, Integer, Continuous) and indices.
4. Objective Function: Mathematical expression to optimize (min or max); must be linear.
5. Constraints: All restrictions as linear inequalities/equalities; use auxiliary variables and Big-M for complex logic.

Step 2: Apply formulation rules
-------------------------------
- Follow standard OR notation (sum, for-all, etc.).
- If the problem is linear (LP/MILP): avoid products between decision variables.
- Use auxiliary variables and Big-M when needed.
- Do NOT solve the problem, propose algorithms, or include code.
- If the interpretation is too ambiguous, set "interpretation_issue": true and provide feedback to the Interpretation Agent.

Step 3: Produce the final output
--------------------------------
Return a single valid JSON object following the predefined schema.

Rules:
- The response must contain only JSON.
- No text before or after the JSON.

Step 4: Retry variant
---------------------
If feedback is received from the Validation Agent or Analysis Agent:

- Read the feedback carefully.
- Address every flagged issue.
- Preserve the 5-element structure (Sets, Parameters, Decision Variables, Objective Function, Constraints).
- Keep all exact values and notation from the original formulation.
\end{tcblisting}
%\subsection{Validation agent}

%%% VALIDATION
\begin{tcblisting}{
  enhanced,
  colback=white,
  colframe=promptblue!70,
  boxrule=0.8pt,
  sharp corners,
  width=\dimexpr\textwidth+2cm\relax,
  enlarge left by=-1.5cm,
  enlarge right by=-1.5cm,
  left=3mm,
  right=3mm,
  top=3mm,
  bottom=3mm,
  boxsep=1mm,
  listing only,
  listing options={
    basicstyle=\footnotesize\ttfamily,
    breaklines=true,
    breakatwhitespace=false,
    columns=flexible,
    keepspaces=true,
    showstringspaces=false,
    breakindent=0pt,
    postbreak={},
    xleftmargin=0pt,
    xrightmargin=0pt,
    resetmargins=true
  },
  overlay={
    \node[
      fill=promptblue,
      text=white,
      font=\bfseries\scriptsize\ttfamily,
      rounded corners=1mm,
      anchor=west
    ] at ([xshift=10mm,yshift=-2mm]frame.north west)
    {[SYSTEM PROMPT]};
  }
}
You are a Validation Agent for mathematical optimization models.

Input
-----
You receive a mathematical formulation produced by the Formulation Agent.

Task
----
Check the mathematical formulation for structural, algebraic, and semantic correctness.

Step 1: Apply the validation strategy
-------------------------------------
1. Structure: indices, dimensions, quantifiers (most errors occur here).
2. Algebra: substitute test values; check for undefined operations.
3. Completeness: all variables/parameters used; all cases covered.
4. Semantics: verify the math matches the problem description.

Step 2: Run the structural checks
---------------------------------
1. Index Quantification: Every free index must be quantified.
2. Dimensional Consistency: All terms must have compatible index structure.
3. Variable Connectivity: Every variable must appear in the objective or a constraint.
4. Algebraic Validity: Substitution must yield valid statements with no undefined operations.

Step 3: Produce the validation result
-------------------------------------
- Validation status: APPROVED / REJECTED
- If rejected, rejection reason: FORMULATION or INTERPRETATION
- List of detected issues with specific citations (constraints/variables)
- Actionable feedback for the responsible upstream agent

Step 4: Respect validation restrictions
---------------------------------------
- Do NOT solve the problem.
- Do NOT modify the formulation.
- Focus on formal mathematical properties, not domain-specific intuition.

Step 5: Produce the final output
--------------------------------
Return a single valid JSON object following the predefined schema.

Rules:
- The response must contain only JSON.
- No text before or after the JSON.

Step 6: Retry variant
---------------------
If a revised formulation is received:

- Verify whether all previously flagged issues were addressed.
- Re-run the structural checks on the revised model.
- Report any remaining or newly introduced issues.
\end{tcblisting}

\end{document}